\documentclass{article}
\usepackage{spconf,amsmath,graphicx}

\usepackage{amssymb}

\title{Mutual Exclusivity Loss for Semi-Supervised Deep Learning}
\name{Mehdi~Sajjadi, Mehran~Javanmardi, and~Tolga~Tasdizen\thanks{This work was supported by NSF IIS-1149299}}
\address{Department of Electrical and Computer Engineering, University of Utah}
\begin{document}
%
\maketitle
\begin{abstract}
In this paper we consider the problem of semi-supervised learning with deep Convolutional Neural Networks (ConvNets). Semi-supervised learning is motivated on the observation that unlabeled data is cheap and can be used to improve the accuracy of classifiers. In this paper we propose an unsupervised regularization term that explicitly forces the classifier's prediction for multiple classes to be mutually-exclusive and effectively guides the decision boundary to lie on the low density space between the manifolds corresponding to different classes of data. Our proposed approach is general and can be used with any backpropagation-based learning method. We show through different experiments that our method can improve the object recognition performance of ConvNets using unlabeled data. 
\end{abstract}

\vspace{-0.5cm}
\section{Introduction}
\vspace{-0.25cm}
Training high accuracy classifiers often requires a very large amount of labeled training data. Recently, ConvNets \cite{lecun3,lecun4} have shown impressive results on many vision tasks including but not limited to classification, detection, localization and scene labeling \cite{overfeat}. However, ConvNets work best when a large amount of labeled data is available for supervised training. For example, the state-of-the-art results for large 1000-category 'ImageNet' \cite{imnet} dataset was significantly improved using ConvNets \cite{imagenet,goognet}. Unfortunately, building large labeled datasets is a costly and time consuming process. 
On the other hand, unlabeled data is easy to obtain. 

Several works have tried to use unlabeled data for training ConvNets. Convolutional deep belief networks \cite{lee2009convolutional} is a generative model for natural images which is based on deep belief networks \cite{hinton2006fast} and trained using unlabeled data. Unlabeled data has also been used for pre-training of convolutional layers in a ConvNet \cite{lecun2010convolutional, jarrett2009best} in an effort to reduce the amount of labeled data required during supervised training. One example is Predictive Sparse
Decomposition (PSD) \cite{kavukcuoglu2010fast} for learning the filter coefficients in the filter bank layer. However, many recent supervised models trained on large datasets usually start from a random initialization of the filter weights which shows that these solutions are not computationally justified. Ladder networks \cite{rasmus2015semi} and region embedding \cite{johnson2015semi} are two more recent examples of semi-supervised learning in ConvNets. 

There are different approaches to semi-supervised learning in general \cite{chapelle2006semi, zhu2005semi}. The classical approaches include self-training, co-training and in general multiview learning \cite{blum1998combining, de1994learning}. In these methods, the strong predictions of a single classifier or multiple classifiers will be added to the training set of the same classifier or other classifiers. Another class of methods for semi-supervised learning is called generative models. There are different methods in this category which are based on Gaussian Mixture Models (GMM) and Hidden Markov Models (HMM) \cite{miller1997mixture}. These generative models generally include the unlabeled data in modeling the probability distribution of the training data and labels. Another approach to semi-supervised learning is Transductive SVM (TSVM) \cite{joachims1999transductive} or S3VM \cite{bennett1999semi}. The goal of these methods is to maximize the classification margin by using the unlabeled data. A large body of semi-supervised approaches are graph-based methods. These methods are generally based on the similarities between labeled and unlabeled samples \cite{blum2001learning, zhu2003semi}. These similarities are encoded in the edges of a graph. In this paper we propose a semi-supervised learning method that makes use of unlabeled data and pushes the decision boundary of Convolutional Neural Networks (CNN) to the less dense areas of decision space and provides better generalization on the test data.
\vspace{-0.5cm}
\subsection{Motivation}
\vspace{-0.2cm}
In many visual classification tasks, it is easy for a human to classify the training samples perfectly; however, the decision boundary is highly nonlinear in the space of pixel intensities. Therefore, we can argue that, the data corresponding to every class lies on a highly nonlinear manifold in the high-dimensional space of pixel intensities and these manifolds don't intersect with each other. An optimal decision boundary lies between the manifolds of different classes where there are no or very few samples. Decision boundaries can be pushed away from training samples by maximizing their margin. Furthermore, it is not necessary to know the class labels of the samples to maximize the margin of a classifier as in TSVMs. However, finding a classifier with a large margin is only possible if the feature set is chosen or found appropriately. For TSVMs the burden is on the kernel of choice. On the other hand, since ConvNets are feature generators without their final fully connected classification layer, if there is a feature space that allows a large margin classifier, they should be capable of finding it in theory. Our argument then is that since object recognition is a relatively easy task for a human, there must be such a feature space that ConvNets can generate with a large margin. Motivated by this argument, we propose a regularization term that uses unlabeled data to encourage the classification layer of a ConvNet to have a large margin. In other words, we propose a regularization term which makes use of unlabeled data and pushes the decision boundary to a less dense area of decision space and forces the set of predictions for a multiclass dataset to be mutually-exclusive.

\vspace{-0.2cm}
\section{Unsupervised Regularization Function}
\begin{figure}[htb]
\begin{minipage}[b]{.48\linewidth}
  \centering
  \centerline{\includegraphics[width=4.0cm]{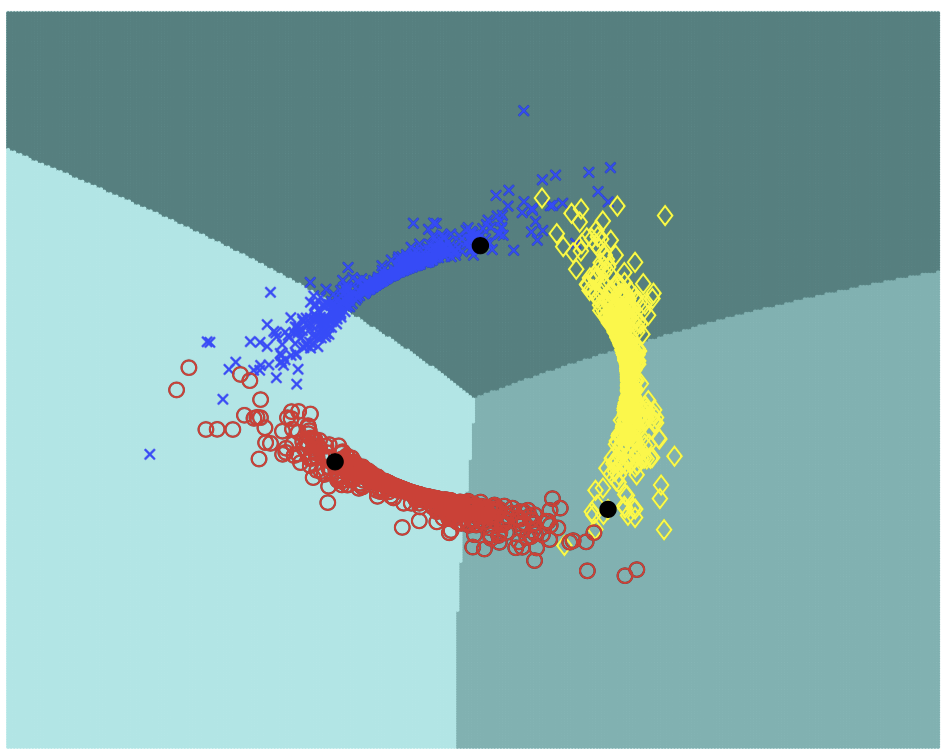}}
  \centerline{(a)}\medskip
\end{minipage}
\hfill
\begin{minipage}[b]{0.48\linewidth}
  \centering
  \centerline{\includegraphics[width=4.0cm]{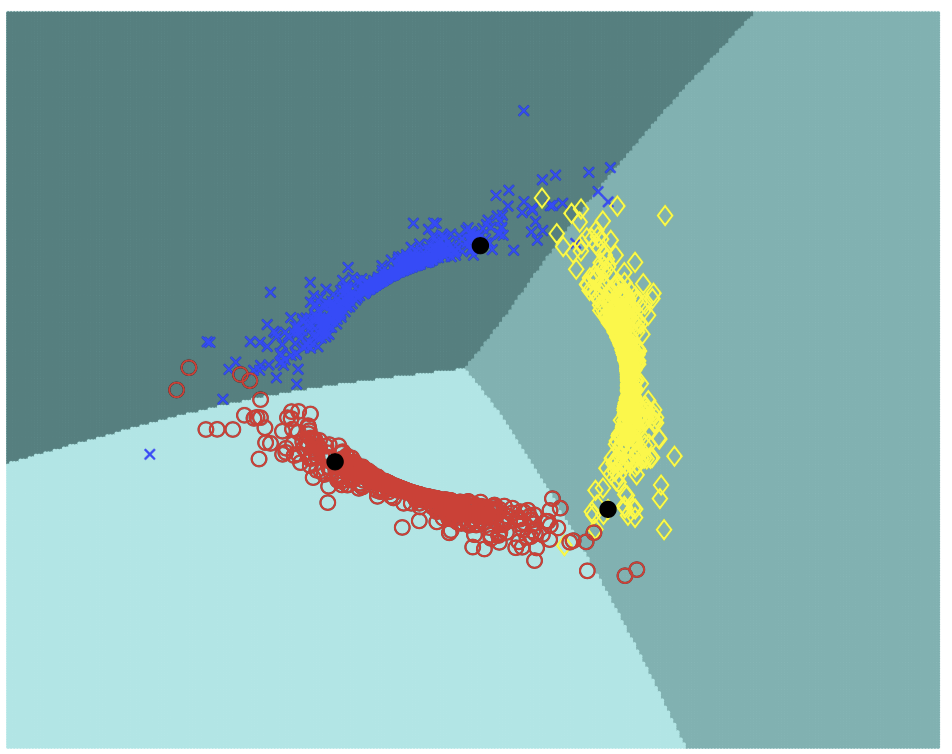}}
  \centerline{(b)}\medskip
\end{minipage}
\vspace{-0.6cm}
\caption{Example showing that unsupervised regularization moves the decision boundary to a less dense area. (a) Without and (b) With unsupervised regularization.}
\label{fig:sy}
\end{figure}

\vspace{-0.2cm}
Let's assume that ${\cal L} = \{(x_i, y_i)\}_{i=1}^{i=L}$ is the set of labeled training data and ${\cal U} = \{(x_i)\}_{i=L+1}^{i=N}$ is the set of unlabeled training data. We also assume that $y_i$ belongs to one of the $K$ classes $\{c_1, c_2, \hdots, c_K \}$. Consider $ \mathbf{f}(\mathbf{w}, \mathbf{x}) $ to be the output vector of a general classifier with learning parameters $\mathbf{w}$ and input vector $\mathbf{x}$. We define $l_{\cal L}(\mathbf{f}(\mathbf{w}, \mathbf{x}_i),y_i)$ to be the loss function defined for the classifier which is calculated based on labeled data of ${\cal L}$. This loss function can be quadratic error, cross-entropy or any other form of loss function. We can assume that in ideal case, the output vector $\mathbf{f}(\mathbf{w}, \mathbf{x})$ is a multi-dimensional binary indicator function $\mathbf{f}:\mathbb{R}^n\rightarrow\mathbb{B}^K$ where $\mathbb{B}=\{0,1\}$ and $n$ is the dimension of input data. If sample $\mathbf{x}_i$ belongs to class $c_k$, then this binary function is in the following form:
\vspace{-0.3cm}
\begin{align}
f_j(\mathbf{w}, \mathbf{x}_i)=\left\{ 
\begin{array}{lr}
1,& j=k\\
0,& j=1 \hdots K,\,\, j \neq k
\end{array}
\right.,
\end{align}
It can be seen that this indicator function can't take any arbitrary vector of the space $\mathbb{B}^K$. In fact it belongs to a very specific subset of this multi-dimensional space which has only one non-zero element. We call this subset $\mathbb{B}^K_s$. We define another binary indicator function $I(\mathbf{f(\mathbf{x},\mathbf{w})})$ which determines if a binary vector $\mathbf{f} \in \mathbb{B}^K$ also belongs to $\mathbb{B}^K_s$ or not. We define this Boolean function $I:\mathbb{B}^K \rightarrow\mathbb{B}$ using disjunction of conjunctions, also known as the disjunctive normal form \cite{Hazewinkel2001}:
\vspace{-0.4cm}
\begin{align}
\label{eq:dnf}
 I(\mathbf{f}) = \bigvee_{j=1}^K{ \left(\bigwedge_{k=1,\, k\neq j}^{K} \lnot f_k \bigwedge f_j \right) }
\end{align}
The output of this indicator function for a valid prediction $\mathbf{f}(\mathbf{x},\mathbf{w})$ of an ideal classifier should be one. This is typically achieved indirectly in a supervised learning setting by one-vs-rest classification which assigns a target value of 1 to the correct class and a target value of 0 to the other classes. However, the labels of the samples are not required if we directly enforce Eq. (\ref{eq:dnf}). In this paper, we enforce this condition in form of a regularization term. To this end, we approximate the binary $I(\mathbf{f})$ with a differentiable function that can be optimized with gradient descent. We replace the conjunction of a set of binary variables $\bigwedge_{i=1}^{K} x_i$ by their product $\prod_{i=1}^K x_i $. We also approximate {\it not} operation of a binary variable $\lnot x_i$ with $1-x_i$. Finally, we substitute the disjunction of the binary variables $\bigvee_{i=1}^{K} x_i$ with their sum $\sum_{i=1}^{K} x_i$. Now relaxing f to be the output of classifiers which are not binary but continuous valued between 0 and 1, $I(\mathbf{f})$ becomes a differentiable function between 0 and 1. By applying the above mentioned approximations, we define the following unsupervised loss function which is calculated using both samples of ${\cal L}$ and ${\cal U}$:
\vspace{-0.3cm}
\begin{align}
\label{eq1}
l_{\cal U}(\mathbf{f}(\mathbf{w}, \mathbf{x}_i)) = -\sum_{j=1}^{K} f_j(\mathbf{w},\mathbf{x}_i) \prod_{k=1,\, k\neq j}^{K} (1-f_k(\mathbf{w},\mathbf{x}_i))
\end{align}
It must be noted that our goal was to maximize $I(\mathbf{f})$. So, we needed to add the minus sign in Eq. (\ref{eq1}) when we define it as a loss function to be minimized. Total loss functions to be minimized is defined as follows:
\vspace{-0.25cm}
\begin{align}
\label{eq11}
l_{\text{tot}} = l_{\cal L} + \lambda l_{\cal U}
\end{align}
This unsupervised loss function $l_{\cal U}$ can be combined with any other loss function and can be used with any backpropagation-based learning. Intuitively, this loss function forces the classifier's prediction to be mutually-exclusive for every class. In addition, it can be observed that this regularization term, forces the decision boundary to be as far as possible from any data sample and as a result it will be placed in a less dense area of decision space. We show this by an example. Figure \ref{fig:sy} shows a synthetic dataset with three classes of diamonds, circles and crosses. Labeled samples are shown with black circles. We trained a simple two layer neural network on this dataset. Decision areas of the neural networks are shown with different colors. Figure \ref{fig:sy} (a) shows the result of the network trained without unsupervised regularization and Figure \ref{fig:sy} (b) is the result of the network with proposed unsupervised regularization. We can see that unsupervised regularization places the decision boundary in a less dense area of space. 
A number of studies \cite{castelli1996relative, o1978normal} show that unlabeled data can be more informative if the classes overlap less. A measure for class overlap is conditional entropy $H(X|Y)$. The empirical conditional entropy can be defined as:
\vspace{-0.25cm}
\begin{align}
\label{eq2}
H(Y|X)=-\frac{1}{N}\sum_{i=1}^{N}\sum_{k=1}^{K}P(c_k | \mathbf{x}_i)\log P(c_k | \mathbf{x}_i)
\end{align}
Note that $P(c_k|\mathbf{x}_i)$ can be estimated with $f_k(\mathbf{w},\mathbf{x}_i)$. It is shown in \cite{grandvalet2004semi} that this entropy minimization can be used to form a regularization term based on unlabeled data. Similar to our proposed method, they try to minimize a loss function which is based on labeled samples and also use a regularization term which is based on Eq. (\ref{eq2}) and calculated using unlabeled data.
However, 
in multiclass problems our regularization term explicitly forces the classifier's prediction for different classes to be mutually-exclusive. We experimentally show that our proposed regularization term generally performs better than entropy minimization which is based on Eq. (\ref{eq2}) on some datasets especially when we have few labeled examples.
\vspace{-0.5cm}
\section{Experiments}
\vspace{-0.3cm}
In this section, we present the results of applying our regularization term for object recognition using ConvNets. We show extensive results on MNIST \cite{lecun4}, CIFAR10 \cite{cifar}, NORB \cite{lecun5} and SVHN \cite{svhn} datasets. We also show some preliminary results on ILSVRC 2012 \cite{imnet} using AlexNet model \cite{imagenet}. In general, we divide the training data of each dataset into two sets and consider one set to be the labeled data ${\cal L}$ and the other set to be unlabeled data ${\cal U}$. In this section, we mainly compare the performance of a ConvNet trained using only labeled data and a ConvNet trained using labeled data and our regularization term calculated using both labeled and unlabeled data. The entropy minimization regularization of \cite{grandvalet2004semi} is also used for training ConvNets in comparison. For all the datasets with the exception of MNIST and ImageNet, we also trained ConvNets with entropy minimization regularization and compared the results with our proposed regularization scheme. It must be noted that the entropy regularization has not been previously used with ConvNets to the best of our knowledge. For every dataset, we train the ConvNet using different ratio of labeled and unlabeled data. In separate experiments, we randomly pick 1\%, 10\%, 50\% and 100\% of training data as labeled set and the rest is reserved for unlabeled set. Then for each setting, we evaluate the improvement obtained by using unlabeled data. We repeat each experiment 5 times for each setting and report the average and standard deviation over error rates of different experiments. During training, the update for model parameters constitute of two parts. The first part is based on labeled data and the second part is from unlabeled data. These two are combined with parameter $\lambda$ according to Eq. (\ref{eq11}). But in our experiments, the labeled set is usually smaller than unlabeled set. So at each epoch we use every labeled sample multiple times in order to compensate the difference in size of labeled and unlabeled datasets. In most of our experiments $\lambda$ is fixed to 1. We experimentally observed that the performance of our regularization method is not overly sensitive to $\lambda$. We incorporated our unsupervised regularization term into cuda-convnet which is a GPU implementation of ConvNet and publicly available at \cite{cudac}. With the exception of MNIST, all other experiments were performed using this GPU implementation. Our setup for all datasets except MNIST and ImageNet consists of 2 convolutional layers followed by two locally connected layers. There are 64 maps in each convolutional layer and 32 maps in each locally connected layer. Filters are 5$\times$5 in convolutional layers and 3$\times$3 in locally connected layers (the same as {\tt 'layers-conv-local-13pct.cfg'} of \cite{cudac} for CIFAR10). We added a fully-connected layer of the size 256 before the output. In all the experiments, we found the number of epochs and learning rates using cross-validation on a small portion of training data and repeat the training on all training data. 
\vspace{-0.5cm}
\begin{table}[ht]
  \caption{Performance comparison on test data for MNIST dataset. error rates: average (\%) $\pm$ std. dev}
  \vspace{-0.15cm}  
  \begin{center}
  \scalebox{0.75}{  
  \begin{tabular}{ c | c | c }
                    &  semi-supervised   &  labeled data only\\ \hline 
           0.13\%   &   14.82 $\pm$ 0.89 & 20.96 $\pm$ 2.19 \\
              1\%   &   1.77  $\pm$ 0.07 & 4.09  $\pm$ 0.22 \\              
             10\%   &   1.28  $\pm$ 0.10 & 1.54  $\pm$ 0.08 \\
             50\%   &   0.87  $\pm$ 0.06 & 0.86  $\pm$ 0.08 \\
             100\%  &   0.79  $\pm$ 0.05 & 0.85  $\pm$ 0.06 \\
  \end{tabular}  
  }
  \end{center}
  \label{tab7}
  \vspace{-1cm}
\end{table}
\begin{table}[ht]
  \caption{Semi-supervised performance comparison on test data for NORB dataset. Error rates: Average (\%) $\pm$ std. dev}
  \vspace{-0.15cm}  
  \begin{center}
  \scalebox{0.75}{  
  \begin{tabular}{ c | c | c | c }
                    & proposed           & labeled data     & entropy        \\ 
                    &  method            &  only            &  minimization  \\ \hline
              1\%   &   19.02 $\pm$ 1.56 & 26.77 $\pm$ 1.60 & 21.12 $\pm$ 0.73 \\
             10\%   &   6.55 $\pm$ 0.37 & 8.42 $\pm$ 0.62   & 7.14  $\pm$ 0.70 \\
             50\%   &   4.58 $\pm$ 0.25 & 4.98 $\pm$ 0.35   & 4.91  $\pm$ 0.21 \\
             100\%  &   4.38 $\pm$ 0.24 & 4.58 $\pm$ 0.31   & 4.24  $\pm$ 0.17 \\
  \end{tabular}
  }  
  \end{center}
  \label{tab3}
  \vspace{-1cm}
\end{table}
\begin{table}[ht]
  \caption{Performance comparison on test data for NORB dataset with fixed labeled set and variable unlabeled set. }
  \vspace{-0.15cm}  
  \begin{center}
  \scalebox{0.75}{    
  \begin{tabular}{ c | c }
                size of unlabeled set    & error rates: average (\%) $\pm$ std. dev  \\ \hline
              labeled data only   &    26.77 $\pm$ 1.60 \\
             25\% of training set  &   21.81 $\pm$ 1.24 \\
             50\% of training set  &   21.48 $\pm$ 2.16 \\
             75\% of training set  &   20.75 $\pm$ 1.21 \\
             100\% of training set  &   19.02 $\pm$ 1.56 \\             
  \end{tabular} 
  } 
  \end{center}
  \label{tab8}
  \vspace{-1cm}
\end{table}
\begin{table}[ht]
  \caption{Semi-supervised performance comparison on test data for SVHN dataset. Error rates: Average (\%) $\pm$ std. dev}
  \vspace{-0.15cm}  
  \begin{center}
  \scalebox{0.75}{  
  \begin{tabular}{ c | c | c | c }
                    & proposed           & labeled data     & entropy        \\ 
                    &  method            &  only            &  minimization  \\ \hline
              1\%   &   15.30 $\pm$ 0.74 & 21.53 $\pm$ 0.53 & 15.23 $\pm$ 0.36 \\
             10\%   &   7.94 $\pm$ 0.15 & 9.88 $\pm$ 0.21   & 7.93 $\pm$ 0.17 \\
             50\%   &   5.65 $\pm$ 0.11 & 6.17 $\pm$ 0.19   & 5.55 $\pm$ 0.22 \\
             100\%  &   4.60 $\pm$ 0.02 & 5.08 $\pm$ 0.07   & 4.68 $\pm$ 0.09 \\
  \end{tabular} 
  } 
  \end{center}
  \label{tab1}
  \vspace{-1cm}
\end{table}
\begin{table}[ht]
  \caption{Performance comparison on test data for SVHN dataset with fixed labeled set and variable unlabeled set.}
  \vspace{-0.15cm}  
  \begin{center}
  \scalebox{0.75}{    
  \begin{tabular}{ c | c }
                size of unlabeled set    &  error rates: average (\%) $\pm$ std. dev  \\ \hline
              labeled data only   &    21.53 $\pm$ 0.53 \\
             25\% of training set  &   16.15 $\pm$ 0.27 \\
             50\% of training set  &   15.55 $\pm$ 0.14 \\
             75\% of training set  &   15.57 $\pm$ 0.23 \\
             100\% of training set  &  15.30 $\pm$ 0.74 \\             
  \end{tabular}  
  }
  \end{center}
  \label{tab9}
  \vspace{-1cm}
\end{table}
\begin{table}[ht]
  \caption{Semi-supervised performance comparison on test data for CIFAR10. Error rates: Average (\%) $\pm$ std. dev}
  \vspace{-0.15cm}  
  \begin{center}
  \scalebox{0.75}{  
  \begin{tabular}{ c | c | c | c }
                    & proposed           & labeled data     & entropy        \\ 
                    &  method            &  only            &  minimization  \\ \hline
              1\%   &   54.48 $\pm$ 0.42 & 57.61 $\pm$ 0.71 & 55.06 $\pm$ 0.65\\
             10\%   &   27.71 $\pm$ 0.32 & 31.48 $\pm$ 0.27 & 27.54 $\pm$ 0.34\\
             50\%   &   16.87 $\pm$ 0.28 & 18.31 $\pm$ 0.31 & 17.02 $\pm$ 0.26\\
             100\%  &   13.67 $\pm$ 0.17 & 14.11 $\pm$ 0.23 & 13.64 $\pm$ 0.23\\
  \end{tabular}  
  }
  \end{center}
  \label{tab5}
  \vspace{-1cm}
\end{table}
\begin{table}[ht]
  \caption{Performance comparison on test data for CIFAR10 dataset with fixed labeled set and variable unlabeled set.}
  \vspace{-0.15cm}
  \begin{center}
  \scalebox{0.75}{
  \begin{tabular}{ c | c }  
                size of unlabeled set    & error rates: average (\%) $\pm$ std. dev  \\ \hline
              labeled data only   &    57.61 $\pm$ 0.71 \\
             25\% of training set  &   55.34 $\pm$ 0.69 \\
             50\% of training set  &   55.09 $\pm$ 0.32 \\
             75\% of training set  &   54.40 $\pm$ 0.65 \\
             100\% of training set  &  54.63 $\pm$ 0.66 \\             
  \end{tabular}  
  }
  \end{center}
  \label{tab10}
  \vspace{-1cm}
\end{table}

\vspace{-0.2cm}
\subsection{MNIST}
\vspace{-0.2cm}
We trained MNIST using a ConvNet with 2 convolutional layers. The first layer uses 7$\times$7 filters and produces 20 maps. The second layer also uses 7$\times$7 
filters but produces 15 maps. A hidden layer with 256 nodes was added before the final layer. No preprocessing was performed on this dataset. We did not use annealing or momentum. For MNIST, we also trained a model with only 80 labeled samples (8 per class). This is equal to 0.13\% of labeled data. The results are given in Table \ref{tab7}. We can see that when there is a small number of labeled data available (0.13\% and 1\% in Table \ref{tab7}), the proposed unsupervised regularization term significantly improves the accuracy.
\vspace{-0.25cm}
\subsection{NORB}
\vspace{-0.2cm}
The training set of NORB contains 10 folds of 29160 images. It is common practice to use only first two folds for training. The test set contains 2 folds totalizing 58320. The original images are 108$\times$108. However, we scaled them down to 
48$\times$48 similar to \cite{mc}. Data translation was used during training. Image translation was obtained by randomly cropping the training images to 44$\times$44. The results are given in Table \ref{tab3}. We can see that in the case with 1\% of labeled data, our supervised term performs better than entropy regularization. The reason is that in this case 1\% of labeled data is not sufficient to guarantee mutual exclusiveness of the predictions of the entropy regularization method. However, our method explicitly forces mutual exclusiveness. In another set of experiments, we fixed the labeled set to be 1\% of total training samples. Then, we increased the size of unlabeled set in 4 steps. We used 25\%, 50\%, 75\% and 100\% of training data as unlabeled set in separate experiments. The results are given in Table \ref{tab8}. It can be seen that by adding more unlabeled data we can improve the performance of classifier.
\vspace{-0.25cm}
\subsection{SVHN}
\vspace{-0.2cm}
SVHN contains around 70000 images for training and more than 500000 easier images for validation. We did not use the validation set at all. The test set contains 26032 images, which are RGB images of size 32 $\times$ 32. Generally, SVHN is a more difficult task than MNIST because of the large variations in the images. We did not do any kind of preprocessing for this dataset. We simply converted the color images to grayscale by removing hue and saturation information. The results are given in Table \ref{tab1}. Similar to NORB, we performed a set of experiments by fixing the size of labeled set and changing the size of unlabeled data. We increased the size of unlabeled set in 4 steps. The results are shown in Table \ref{tab9}. Here again, we observe that by increasing the size of unlabeled data we can actually improve the classification performance.
\subsection{CIFAR10}
\vspace{-0.2cm}
We augmented the training data using image translations, which is done by taking 24$\times$24 cropped versions of the original images at random locations. A common preprocessing for this dataset
is to subtract the per pixel mean of the training set from every image. The results are given in Table \ref{tab5}.
Similar to NORB and SVHN, we fixed the labeled set at 1\% of training data and increased the size of unlabeled set in 4 steps. The results are given in Table \ref{tab10}. We can see that the performance keeps improving as we add more unlabeled data.

\vspace{-0.3cm}
\subsection{ImageNet}
\vspace{-0.2cm}
We performed preliminary experiments with ILSVRC 2012 which has 1000 classes. We randomly picked 10\% of each class from training data as labeled set and the rest was used for unlabeled set. We applied our regularization term to AlexNet model \cite{imagenet}. Using our method we achieved an error rate of 42.90\%. If we don't use the regularization term the error rate is 45.63\%. This shows that our model can be effective even when we have large number of classes.

\vspace{-0.3cm}
\section{Discussion}
\vspace{-0.2cm}
In all of the experiments we observed performance improvement by using unsupervised regularization. 
Based on our experiments we can see that for the cases with very few labeled samples, the advantage of using unsupervised regularization term is more significant when the classification task is simpler. For example, CIFAR10 is a more challenging dataset compared to MNIST, NORB and SVHN and benefits less from using unsupervised term. In simpler tasks, ConvNet is able to create a feature space with less dense areas and provides better discriminative features for the classifier of final layer. This means that even for more challenging tasks, if more unsupervised training data becomes available, then large ConvNets can be trained which might be able to create the feature spaces that will have less dense areas and larger margins.
\vspace{-0.5cm}
\section{Conclusion}
\vspace{-0.2cm}
We introduced an unsupervised regularization term that forces a classifier predictions to be mutually-exclusive for different classes and moves the decision boundary to a less dense area of decision space. We showed that our method can be applied successfully to ConvNets to improve the classification accuracy using both labeled and unlabeled data. We showed that it is possible to improve classification accuracy by adding more unlabeled data. We also showed that entropy regularization can be applied to ConvNets successfully. 


%



\bibliographystyle{IEEEbib}
{\small
\bibliography{refs}}

\begin{thebibliography}{10}

\bibitem{lecun3}
B~Boser Le~Cun, John~S Denker, D~Henderson, Richard~E Howard, W~Hubbard, and
  Lawrence~D Jackel,
\newblock ``Handwritten digit recognition with a back-propagation network,''
\newblock in {\em Advances in neural information processing systems}. Citeseer,
  1990.

\bibitem{lecun4}
Yann LeCun, L{\'e}on Bottou, Yoshua Bengio, and Patrick Haffner,
\newblock ``Gradient-based learning applied to document recognition,''
\newblock {\em Proceedings of the IEEE}, vol. 86, no. 11, pp. 2278--2324, 1998.

\bibitem{overfeat}
Pierre Sermanet, David Eigen, Xiang Zhang, Micha{\"e}l Mathieu, Rob Fergus, and
  Yann LeCun,
\newblock ``Overfeat: Integrated recognition, localization and detection using
  convolutional networks,''
\newblock {\em arXiv preprint arXiv:1312.6229}, 2013.

\bibitem{imnet}
Alex Berg, Jia Deng, and L~Fei-Fei,
\newblock ``Large scale visual recognition challenge 2010,'' 2010.

\bibitem{imagenet}
Alex Krizhevsky, Ilya Sutskever, and Geoffrey~E Hinton,
\newblock ``Imagenet classification with deep convolutional neural networks,''
\newblock in {\em Advances in neural information processing systems}, 2012, pp.
  1097--1105.

\bibitem{goognet}
Christian Szegedy, Wei Liu, Yangqing Jia, Pierre Sermanet, Scott Reed, Dragomir
  Anguelov, Dumitru Erhan, Vincent Vanhoucke, and Andrew Rabinovich,
\newblock ``Going deeper with convolutions,''
\newblock {\em arXiv preprint arXiv:1409.4842}, 2014.

\bibitem{lee2009convolutional}
Honglak Lee, Roger Grosse, Rajesh Ranganath, and Andrew~Y Ng,
\newblock ``Convolutional deep belief networks for scalable unsupervised
  learning of hierarchical representations,''
\newblock in {\em Proceedings of the 26th Annual International Conference on
  Machine Learning}. ACM, 2009, pp. 609--616.

\bibitem{hinton2006fast}
Geoffrey~E Hinton, Simon Osindero, and Yee-Whye Teh,
\newblock ``A fast learning algorithm for deep belief nets,''
\newblock {\em Neural computation}, vol. 18, no. 7, pp. 1527--1554, 2006.

\bibitem{lecun2010convolutional}
Yann LeCun, Koray Kavukcuoglu, and Cl{\'e}ment Farabet,
\newblock ``Convolutional networks and applications in vision,''
\newblock in {\em Circuits and Systems (ISCAS), Proceedings of 2010 IEEE
  International Symposium on}. IEEE, 2010, pp. 253--256.

\bibitem{jarrett2009best}
Kevin Jarrett, Koray Kavukcuoglu, Marc'Aurelio Ranzato, and Yann LeCun,
\newblock ``What is the best multi-stage architecture for object
  recognition?,''
\newblock in {\em Computer Vision, 2009 IEEE 12th International Conference on}.
  IEEE, 2009, pp. 2146--2153.

\bibitem{kavukcuoglu2010fast}
Koray Kavukcuoglu, Marc'Aurelio Ranzato, and Yann LeCun,
\newblock ``Fast inference in sparse coding algorithms with applications to
  object recognition,''
\newblock {\em arXiv preprint arXiv:1010.3467}, 2010.

\bibitem{rasmus2015semi}
Antti Rasmus, Mathias Berglund, Mikko Honkala, Harri Valpola, and Tapani Raiko,
\newblock ``Semi-supervised learning with ladder networks,''
\newblock in {\em Advances in Neural Information Processing Systems}, 2015, pp.
  3532--3540.

\bibitem{johnson2015semi}
Rie Johnson and Tong Zhang,
\newblock ``Semi-supervised convolutional neural networks for text
  categorization via region embedding,''
\newblock in {\em Advances in Neural Information Processing Systems}, 2015, pp.
  919--927.

\bibitem{chapelle2006semi}
Olivier Chapelle, Bernhard Sch{\"o}lkopf, Alexander Zien, et~al.,
\newblock ``Semi-supervised learning,''
\newblock 2006.

\bibitem{zhu2005semi}
Xiaojin Zhu,
\newblock ``Semi-supervised learning literature survey,''
\newblock 2005.

\bibitem{blum1998combining}
Avrim Blum and Tom Mitchell,
\newblock ``Combining labeled and unlabeled data with co-training,''
\newblock in {\em Proceedings of the eleventh annual conference on
  Computational learning theory}. ACM, 1998, pp. 92--100.

\bibitem{de1994learning}
Virginia~R de~Sa,
\newblock ``Learning classification with unlabeled data,''
\newblock in {\em Advances in neural information processing systems}, 1994, pp.
  112--119.

\bibitem{miller1997mixture}
David~J Miller and Hasan~S Uyar,
\newblock ``A mixture of experts classifier with learning based on both
  labelled and unlabelled data,''
\newblock in {\em Advances in neural information processing systems}, 1997, pp.
  571--577.

\bibitem{joachims1999transductive}
Thorsten Joachims,
\newblock ``Transductive inference for text classification using support vector
  machines,''
\newblock in {\em ICML}, 1999, vol.~99, pp. 200--209.

\bibitem{bennett1999semi}
Kristin Bennett, Ayhan Demiriz, et~al.,
\newblock ``Semi-supervised support vector machines,''
\newblock {\em Advances in Neural Information processing systems}, pp.
  368--374, 1999.

\bibitem{blum2001learning}
Avrim Blum and Shuchi Chawla,
\newblock ``Learning from labeled and unlabeled data using graph mincuts,''
\newblock 2001.

\bibitem{zhu2003semi}
Xiaojin Zhu, Zoubin Ghahramani, John Lafferty, et~al.,
\newblock ``Semi-supervised learning using gaussian fields and harmonic
  functions,''
\newblock in {\em ICML}, 2003, vol.~3, pp. 912--919.

\bibitem{Hazewinkel2001}
M~Hazewinkel, Ed.,
\newblock {\em Encylopedia of Mathematics},
\newblock Springer, 2001.

\bibitem{castelli1996relative}
Vittori Castelli and Thomas~M Cover,
\newblock ``The relative value of labeled and unlabeled samples in pattern
  recognition with an unknown mixing parameter,''
\newblock {\em Information Theory, IEEE Transactions on}, vol. 42, no. 6, pp.
  2102--2117, 1996.

\bibitem{o1978normal}
Terence~J O'neill,
\newblock ``Normal discrimination with unclassified observations,''
\newblock {\em Journal of the American Statistical Association}, vol. 73, no.
  364, pp. 821--826, 1978.

\bibitem{grandvalet2004semi}
Yves Grandvalet and Yoshua Bengio,
\newblock ``Semi-supervised learning by entropy minimization,''
\newblock in {\em Advances in neural information processing systems}, 2004, pp.
  529--536.

\bibitem{cifar}
Alex Krizhevsky and Geoffrey Hinton,
\newblock ``Learning multiple layers of features from tiny images,'' 2009.

\bibitem{lecun5}
Yann LeCun, Fu~Jie Huang, and Leon Bottou,
\newblock ``Learning methods for generic object recognition with invariance to
  pose and lighting,''
\newblock in {\em Computer Vision and Pattern Recognition, 2004. CVPR 2004.
  Proceedings of the 2004 IEEE Computer Society Conference on}. IEEE, 2004,
  vol.~2, pp. II--97.

\bibitem{svhn}
Yuval Netzer, Tao Wang, Adam Coates, Alessandro Bissacco, Bo~Wu, and Andrew~Y
  Ng,
\newblock ``Reading digits in natural images with unsupervised feature
  learning,''
\newblock in {\em NIPS workshop on deep learning and unsupervised feature
  learning}. Granada, Spain, 2011, vol. 2011, p.~5.

\bibitem{cudac}
A~Krizhevskey,
\newblock ``Cuda-convnet,'' code.google.com/p/cuda-convnet, 2014.

\bibitem{mc}
Dan Ciresan, Ueli Meier, and J{\"u}rgen Schmidhuber,
\newblock ``Multi-column deep neural networks for image classification,''
\newblock in {\em Computer Vision and Pattern Recognition (CVPR), 2012 IEEE
  Conference on}. IEEE, 2012, pp. 3642--3649.

\end{thebibliography}


\end{document}